\documentclass[lettersize,journal]{IEEEtran}
\usepackage{amsmath,amsfonts}
\usepackage{algorithmic}
\usepackage{algorithm}
\usepackage{array}
\usepackage[caption=false,font=normalsize,labelfont=sf,textfont=sf]{subfig}
\usepackage{textcomp}
\usepackage{stfloats}
\usepackage{url}
\usepackage{verbatim}
\usepackage{graphicx}
\usepackage{cite}
\usepackage{xcolor}
\usepackage{hyperref}
\usepackage{multirow}
\usepackage{makecell}  
\usepackage{nicematrix}
\usepackage{caption}
\usepackage[table,dvipsnames,x11names]{xcolor}
\usepackage{float}
\usepackage{booktabs}
\usepackage{graphicx}
\usepackage{siunitx}
\usepackage{amssymb}
\usepackage{threeparttable}
\usepackage{booktabs, siunitx, tabularx, xcolor}
\usepackage{siunitx}
\usepackage[table]{xcolor}
\usepackage[utf8]{inputenc}
\usepackage[T1]{fontenc}
\usepackage{newunicodechar}

\newunicodechar{，}{,}
\newunicodechar{。}{.}
\newunicodechar{：}{:}
\newunicodechar{；}{;}
\newunicodechar{！}{!}
\newunicodechar{？}{?}
\newunicodechar{（}{(}
\newunicodechar{）}{)}
\newunicodechar{、}{,}
\newunicodechar{“}{''}
\newunicodechar{”}{''}
\newunicodechar{—}{-}
\newunicodechar{…}{\ldots}

\begin{document}

\title{VRS-UIE: Value-Driven Reordering Scanning \\ for Underwater
 Image Enhancement}

\author{
        Kui~Jiang,~\IEEEmembership{Member~IEEE}, 
        Yan~Luo, 
        Junjun~Jiang,~\IEEEmembership{Senior Member~IEEE}, 
        Ke~Gu,~\IEEEmembership{Senior Member~IEEE},
        Nan~Ma,~\IEEEmembership{Senior Member~IEEE},
        Xianming~Liu,~\IEEEmembership{Senior Member~IEEE}\\

\thanks{This research was financially supported by the National Natural Science Foundation of China (62501189), Natural Science Foundation of Heilongjiang Province of China for Excellent Youth Project (YQ2024F006), Open Research Fund from Guangdong Laboratory of Artificial Intelligence and Digital Economy (SZ) (GML-KF-24-09).}
\thanks{K. Jiang, Y. Luo, J. Jiang and X. Liu are with the School of Computer Science and Technology, Harbin Institute of Technology, China, 150001 (jiangkui@hit.edu.cn, luoyan@stu.hit.edu.cn, jiangjunjun@hit.edu.cn, csxm@hit.edu.cn).}
\thanks{K. Gu and N. Ma are with the Faculty of Information Technology, Beijing University of Technology, China, 100124 (guke@bjut.edu.cn, manan123@bjut.edu.cn).}
}

\markboth{}
{Jiang \MakeLowercase{\textit{et al.}}: VRS-UIE: Value-Driven Reordering Scanning for Underwater Image Enhancement}


\maketitle

\begin{abstract}
State Space Models (SSMs) have emerged as a promising backbone for vision tasks due to their linear complexity and global receptive field. However, in the context of Underwater Image Enhancement (UIE), the standard sequential scanning mechanism is fundamentally challenged by the unique statistical distribution characteristics of underwater scenes. The predominance of large-portion, homogeneous but useless oceanic backgrounds can dilute the feature representation responses of sparse yet valuable targets, thereby impeding effective state propagation and compromising the model's ability to preserve both local semantics and global structure. 
To address this limitation, we propose a novel Value-Driven Reordering Scanning framework for UIE, termed VRS-UIE. Its core innovation is a Multi-Granularity Value Guidance Learning (MVGL) module that generates a pixel-aligned value map to dynamically reorder the SSM's scanning sequence. This prioritizes informative regions to facilitate the long-range state propagation of salient features. Building upon the MVGL, we design a Mamba-Conv Mixer (MCM) block that synergistically integrates priority-driven global sequencing with dynamically adjusted local convolutions, thereby effectively modeling both large-portion oceanic backgrounds and high-value semantic targets. A Cross-Feature Bridge (CFB) further refines multi-level feature fusion. Extensive experiments demonstrate that our VRS-UIE framework sets a new state-of-the-art, delivering superior enhancement performance (surpassing WMamba by 0.89 dB on average) by effectively suppressing water bias and preserving structural and color fidelity. Furthermore, by incorporating efficient convolutional operators and resolution rescaling, we construct a light-weight yet effective scheme, VRS-UIE-S, suitable for real-time UIE applications.
\end{abstract}



\section{Introduction}

\IEEEPARstart{U}{nderwater} image enhancement (UIE) aims to restore the visual quality of scenes degraded by light absorption and scattering~\cite{Waternet,Ucolor}.
These degradations manifest themselves as low contrast, severe color distortion, and blurred textures, which alter human visual perception and hinder downstream computer vision tasks such as autonomous navigation, marine resource exploration, and biological monitoring~\cite{IVP,uie_od}. As a fundamental preprocessing step, UIE is critical to ensuring the reliability of underwater vision systems.

Prior to the advent of deep learning, conventional methods~\cite{6247661,5642311,7574330} rely on hand-crafted priors. These approaches often exhibit limited generalization in diverse underwater environments due to oversimplified physical assumptions. Recent deep learning-based methods~\cite{chen2021underwater,10.1016/j.knosys.2022.110041} have demonstrated improved performance. However, they remain constrained: Convolutional Neural Networks (CNNs) are limited by their effective receptive fields~\cite{UWCNN,Ucolor}, while vision Transformers capture long-range dependencies at a high computational cost, especially for high-resolution images~\cite{10129222}. 


To balance global modeling with computational efficiency, recent research has explored Mamba-based state space models (SSMs)~\cite{mamba,SSS} as alternative backbones, benefiting from their linear complexity and strong global representation capabilities. This has spurred innovation in architectures and training strategies, including those focused on directional coverage~\cite{VMamba,Vision_mamba}, receptive scope~\cite{omamba,tinyvim}, and scanning granularity~\cite{LocalMamba,MSmamba,GroupMamba}. 
Despite this progress, existing SSM-based technologies struggle to preserve both local object semantics and global structure when applied to UIE. This shortcoming stems from the inherent limitations of the one-dimensional sequential scanning process and the unique distribution characteristics of underwater images.

Specifically, in typical underwater scenes, the homogeneous oceanic backgrounds occupies a large portion of the image. This counterpart provides few informative visual cues (\emph{e.g.}, object boundaries, salient foreground, texture) and dilutes the feature representation of valuable targets, such as marine organisms. Furthermore, because the state update in an SSM depends on all previously visited positions in the scanning sequence, the feature response for sparse but valuable targets is barely activated during long-range propagation, which easily interferes with the prevalent and useless background. 

Motivated by these insights, we propose a value-driven reordering strategy to improve sequential scanning by prioritizing informative and valuable content. This facilitates long-range state propagation for valuable components by generating purer and more efficient states, thereby enabling high-quality color inference and structure reconstruction. 
Building on this principle, we design a Multi-Granularity Value Guidance Learning (MVGL) module that provides the value map to adaptively guide the holistic content reordering by aligning an implicit sampling frequency knowledge with an explicit DINO-based valuable prior. 
The implicit branch within MVGL dynamically calculates sampling offset frequencies based on structural affinity, similarity, and local semantics. The frequencies refer to the activation count of the current pixel by others, reflecting the position distribution of globally critical components. 
The explicit branch produces a token-level valuable map derived from DINOv2 features based on the L2-norm across channel dimensions. This map guides the implicit branch during training by computing a multi-granularity KL divergence loss, which encourages the model to focus on valuable regions and mitigates uncertainty caused by diverse distribution characteristics of underwater scenes. 

Based on MVGL, we develop a new Mamba–Conv Mixer (MCM) block to explore contextual information from global regions based on dual-path cooperation between semantic relevance and structural significance. The core Mixer comprises two complementary paths. 
The reordering Mamba path processes the reordered features to achieve value-driven state propagation, anchored to visually salient regions. 
The dynamic convolution path mitigates global focus bias by adaptively reshaping convolutional responses through dynamic kernel adjustments. This improves local structurally semantic representation and adaptability, complementing global sequence modeling. 
The dual-branch design is greatly compatible with the distinct characteristics of underwater scenes: oceanic backgrounds exhibit large distribution areas but lower statistical frequency due to the vote splitting effect, while marine organisms (\emph{e.g.}, fish, corals) represent high statistical frequency and local semantic. 
To further refine feature integration, we introduce a Cross-Feature Bridge (CFB) that facilitates multi-level fusion among encoder and decoder through learnable hybrid attention, ensuring robust hierarchical feature extraction and representation.
In general, these components form our proposed Value-Driven Reordering Scanning framework (VRS-UIE) for effective underwater image enhancement.

\begin{figure*}
    \centering
    \includegraphics[width=1\linewidth]{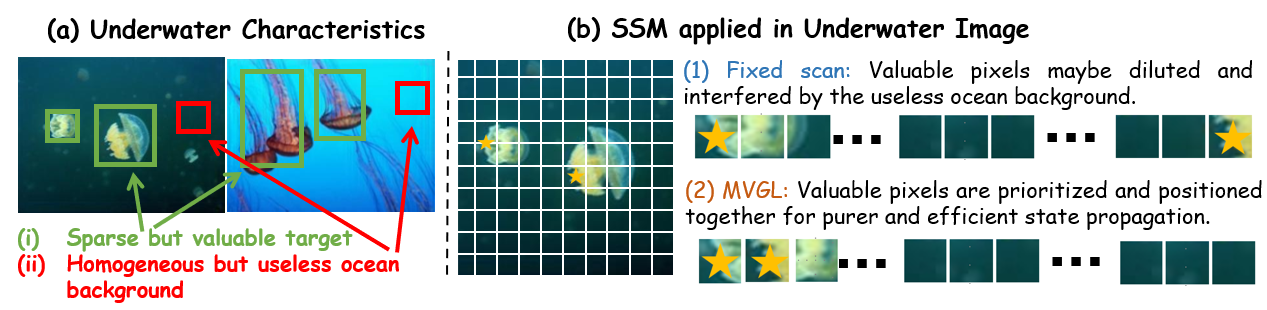}
    \caption{(a) Distribution characteristics of underwater images: 
    informative regions are sparse but valuable (green boxes), while the oceanic backgrounds is large-portion, homogeneous and useless (red boxes). (b) When an SSM uses a (1) fixed scan , valuable pixels are dispersed along the sequence and their 
    feature responses are barely activated and diluted by 
    useless oceanic backgrounds during long-range propagation. (2) MVGL (ours) applies a value-priority schedule: a learnable value map reorders tokens so high-value pixels are \textbf{prioritized and positioned together}, enabling purer and more efficient state propagation. 
}
    \label{fig:motivation}
\end{figure*}

The main contributions are summarized as follows.

\begin{itemize}
    
\item
The paper introduces a Value-Driven Reordering Scanning framework (VRS-UIE) for underwater image enhancement. Instead of scanning pixels in a fixed order, it prioritizes regions with a higher semantic value or structural significance. This ensures that long-range context is propagated from the most informative parts of the image first, leading to more effective color and structure reconstruction.
    

\item  
This paper presents a Multi-Granularity Value Guidance Learning (MVGL) 
module to learn the priority ranking with multi-granularity guidance, which facilitates the value map generation while enhancing adaptability to diverse distribution of underwater scene.  

\item
The paper advises a dual-path Mamaba-Conv Mixer (MCM) block designed to handle the distinct characteristics of underwater scenes, which harmonizes the value-reordered global knowledge propagation and local semantic representation. 

    
\end{itemize}

\section{Related Work}
\subsection{Underwater Image Enhancement}
Existing underwater image enhancement (UIE) methods are broadly categorized into physics-based and learning-based approaches. Early model-driven techniques, such as UDCP~\cite{6755982} and IBLA~\cite{7840002}, operate by estimating underwater optical parameters like background light and a wavelength-dependent attenuation model. 
However, these methods rely on idealized assumptions and often perform poorly 
in complex, non-uniform underwater environments. Learning-based methods, such as UWCNN, learn degraded-to-clear image mappings directly via CNNs but frequently neglect underlying physical priors, leading to suboptimal color restoration. 
Subsequent two-stage frameworks like CLCE-Net~\cite{YIN2022109997} decouple color correction and deblurring; however, they often lack explicit physical constraints. While Water-Net~\cite{C60} introduces physical parameters, it faces optimization instability due to tightly coupled parameter interactions.

\subsection{State Space Models }
Mamba~\cite{Gu2023MambaLS} represents a breakthrough in selective state space models (SSMs), introducing a data-dependent mechanism for state transitions with linear computational complexity. Its success in natural language processing has spurred adaptations to computer vision tasks, including 3D analysis, medical imaging, and low-level vision applications. Recent work in image restoration demonstrates Mamba's potential: MatIR~\cite{Wen2025MatIRAH} develops a hybrid Mamba–Transformer architecture with interleaved modules and multi-path SSM processing; FMSR~\cite{xiao2024frequency} combines frequency-domain priors with Mamba for remote sensing image enhancement; and Retinex-RAW-Mamba~\cite{Chen2024RetinexRAWMambaBD} integrates directional scanning with Retinex decomposition for RAW image processing. 
Despite these advances, conventional Mamba architectures face two key limitations in the context of UIE: 
1) Flattening the image into a 1D sequence can exacerbate the difficulty of modeling the sparse and critical salient regions typical in underwater scenes. 
2) The characteristic of SSMs, where the prefix sequence guides the recovery of subsequent content, is often overlooked.
To address these challenges, we introduce a reordering strategy via a Multi-Granularity Value Guidance Learning (MVGL) module. 
This approach prioritizes the state propagation of valuable regions, thereby providing purer and more efficient states to facilitate more effective color inference and structure reconstruction.

\begin{figure*}[htbp]
    \centering
    \includegraphics[width=0.985\linewidth]{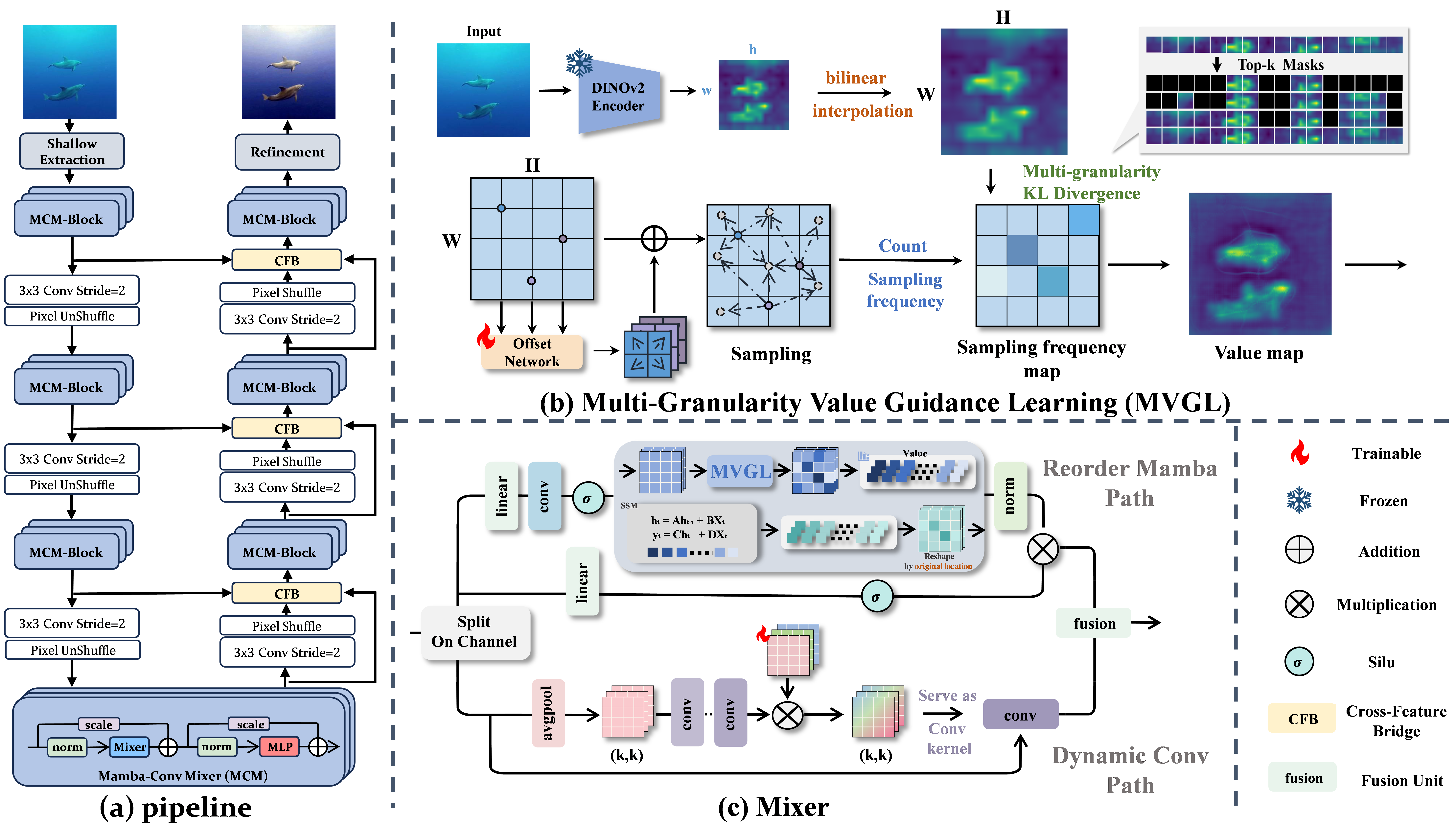}
    \caption{An overview of the proposed Value-Driven Reordering Scanning framework (VRS-UIE). (a) The overall pipeline, which is based on a U-Net architecture with MCM blocks at each stage. 
    (b) The Multi-Granularity Value Guidance Learning (MVGL) module.  
    (c) The Mixer module within the MCM block, featuring a dual-path design with a reordering Mamba path and a dynamic convolution, 
    which are designed to explore contextual information from regions based on semantic relevance or structural significance. 
    }\vspace{-4mm}
    \label{fig:pipeline}
\end{figure*}

\subsection{Dynamic Convolution}
While CNNs are fundamental to image enhancement, their static convolutional kernels, featuring fixed weights and sampling location, 
struggle to address the spatial heterogeneity of underwater degradation. 
Dynamic convolution addresses this limitation through two primary strategies. \textbf{Spatial-adaptive convolution:} Deformable Convolutional Networks (DCN) pioneer adaptive sampling by learning spatial offsets. 
Subsequent work, such as DAT~\cite{xia2022vision}, integrates attention mechanisms to improve spatial selectivity, although these methods typically retain fixed kernel weights. 
\textbf{Weight-adaptive convolution:} CondConv~\cite{yang2019condconv} generates dynamic weights based on conditional parameters, while DynamicConv~\cite{chen2020dynamic} employs a gating mechanism for expert kernel selection. DyNet~\cite{zhang2020dynet} augments this approach with channel attention, but incurs a high parameter cost due to full-channel modulation.
To achieve a more favorable balance between flexibility and efficiency, we propose a grouped dynamic convolution that synthesizes content-aware kernel weights through channel-wise grouping and global pooling. This design aligns the input dependencies with the Mamba branch, enabling unified global-local modeling.

\section{Methodology}
Our network follows a U-Net design and is instantiated with Mamba–Conv Mixer (MCM) blocks to reduce color bias and strengthen local structural detail, improving color fidelity and texture preservation.
The core Mixer employs a dual-branch configuration : 
(i) a Reordering Mamba path using the order learned by MVGL to perform value-driven scanning, thereby enabling the capture of global and content-aware dependencies;
and (ii) a dynamic convolution path \cite{TransXNet} that generates a dynamic kernel from the current input and applies it uniformly to restore 2D inductive bias, suppress long-range accumulation noise in 1D Mamba propagation, and enhance edges and textures in diverse underwater scenes, complementing global modeling.
Furthermore, a Cross-Feature Bridge (CFB) is integrated into the U-Net skip connections to fuse multi-level encoder–decoder features via a learnable hybrid attention gating mechanism, thereby enhancing color bias correction and detail enhancement.

\subsection{Mamba–Conv Mixer (MCM)}

As illustrated in Figure \ref{fig:pipeline}(a), our U-Net backbone is built from cascaded MCM blocks, adopting the Transformer-style pre-norm residual layout (LayerNorm~$\rightarrow$~Mixer~$\rightarrow$~LayerNorm~$\rightarrow$~MLP with residual connections).

The core Mixer module consists of two structurally complementary branches : reordering Mamba and dynamic convolution. As shown in Figure \ref{fig:pipeline}(c), we first divide the feature map $X \in \mathbb{R}^{H \times W \times C}$ along the channel into two parts notated as $X_1 \in \mathbb{R}^{H \times W \times C_1}$ and $X_2 \in \mathbb{R}^{H \times W \times C_2}$, where $C_1=C_2=\frac{C}{2}$. Then we feed $X_1$ and $X_2$ into the reordering Mamba path and the dynamic convolution path.

\noindent \textbf{Reordering Mamba.} 
Unlike fixed-scan Mamba, our Reordering Mamba performs a value-priority scan. Specifically, we generate a learnable value map $V$ to reorder the sequence so that high-value pixels are visited early and clustered in a compact prefix, producing cleaner and more efficient state propagation.

To improve the robustness of $V$, we introduce Multi-Granularity Value Guidance Learning (MVGL). 
As shown in Figure \ref{fig:pipeline} (b), an implicit branch predicts sampling offsets and counts the cumulative quantity to produce a deformable sampling frequency map $S$. 
During training, an explicit DINOv2 prior $D$ is used to guide the prediction of $S$ via a multi-granularity KL loss. This scheme introduces a set of masks with different Top-$k$ ratios to distill the value cues from $D$, which encourages the model to focus on valuable regions and mitigate the uncertainty from distribution variation.
The resulting value map $V$ is then used as the ranking score for the reordering of the scan path.

\textit{(i) Explicit Guided valuable map $D$.} We compute an explicit token-level valuable prior by taking the L2 norm over the channel dimension of DINOv2 features.
  $  D \in \mathbb{R}^{h\times w}(h=w=16),$
which is globally informative, yet localized coarse. $D$ provides guidance during training, which is detached from the gradients and \textbf{not invoked in inference}.

\textit{(ii) Implicit Sampling frequency map $S$.} 
We view the implicit branch as a self-similarity sampling mechanism: each position predicts multiple offsets via a shared offset predictor with strong local affinity. Under this shared predictor, structurally similar positions tend to sample analogous regions—high-value prospects engage in mutual cross-sampling, whereas low-information regions exhibit divergence in their sampling behavior. By aggregating sampling frequencies across positions, we construct a sampling frequency map S, which is also guided by explicit branches.

The specific construction method of the cumulative sampling statistical distribution $S$ is as follows. For each query position $p$, a lightweight network is employed to predict sampling offsets corresponding to a set of spatial displacements ($k^2$).
The aforementioned procedures are defined as 
\begin{equation}
    \Delta^n(p) = f_{1 \times 1}(\text{GELU}(f_{\text{depth}}(X_1))) \quad 
    n = 1,...,k^2 ,\\
\end{equation}  
where $f_{depth}$ refers to a $3\times3$ depthwise convolution and $f_{1\times1}$ denotes a $1\times1$ convolution. 

With the predicted offsets $\Delta^n(p)$ and the original coordinates, we 
generate the new sampling locations $\hat{p}^n$.
In this way, we can calculate the sampling frequency for each position $p$ to output the sampling frequency map $S$, depicted as 
\begin{align}
\hat{p}^n = p_{\text{base}} + \Delta^ n (p) , \\
    S(p) =  \sum_{n=1}^{k^2} \mathbb{I}~(p \in \hat{p}^n) . 
\end{align}

\textit{(iii) Multi-granularity guidance.}
To eliminate the resolution gap, we upsample the $D \in \mathbb{R}^{16 \times 16}$ to generate $D^\uparrow$, which is of the same size as $S \in \mathbb{R}^{H \times W}$. 
With temperature T, we normalize
\begin{align}
    p = softmax(Vec(S)/T) ,\\
    q = softmax(Vec(D^\uparrow)/T) .
\end{align}
We use a linear annealing schedule T(t) for the first half of training: $T(0) = 1.0 $→$ T(T_{half}) = 0.1$, and then keep T = 0.1 thereafter. A larger $T$ produces softer distributions and smaller gradients, which stabilizes early training and avoids over-committing to the value prior. We gradually lowering $T$ sharpens $p$ and $q$, strengthening the guidance on high-value regions in later stages.

Since the distribution characteristic varies widely across underwater scenes, we therefore adopt a multi-granularity KL divergence loss to guide the alignment between $D$ and sampling frequency map $S$. Specifically, we extract components from the guided value map $D$ and the sampling frequency map $S$ via a set of masks $M^{(i)}$ with diverse Top-$k$ ratios $\{r_i\}$ ($25/50/75/100\% $) on q. 
Within each mask, 
the KL loss is given by

\begin{align}
    p^{(i)} = \frac{p \odot M^{(i)}}{|| p \odot M^{(i)} ||_1} ,~~~
    q^{(i)} = \frac{q \odot M^{(i)}}{|| q \odot M^{(i)} ||_1} ,\\
    KL(q^{(i)} || p^{(i)}) = \sum_j q^{(i)}_j log \frac{q^{(i)}_j} {p^{(i)}_j + \epsilon} , \\
    L_{top-k} = \sum_i KL(q^{(i)} || p^{(i)}),
\end{align}
where $\epsilon = 1e^{-8}$ used to avoid calculation errors. Guided by the explicit value prior $D$ via multi-granularity KL, the sampling frequency map $S$ is turned into the final value map $V$. We then use $V$ as the 
prior guidance to reorder the scan path in SSM.

\textit{(iv) Reorder by value map.} 
To reorder our feature map according to the value map, we first flatten the value map \(V \in \mathbb{R}^{H \times W}\) into a vector \(v \in \mathbb{R}^{N}\), where \(N = HW\). We then obtain a descending permutation \(\pi = \mathrm{argsort}_{\downarrow}(v)\). This permutation is used to gather the feature map from \(X \in \mathbb{R}^{H \times W \times C}\) to the sequence \(X^{\mathrm{seq}} \in \mathbb{R}^{N \times C}\), where each element is defined as \(X^{\mathrm{seq}}_{t} = X_{\pi(t),:}\).

The structured state space model (SSM) processes this sequence in the sorted order. After processing, the output sequence is scattered back to the original spatial arrangement using inverse permutation $\pi ^{-1}$. Throughout this reordering process, the SSM equations remain unchanged:
\begin{align}
     h_t = A h_{t-1} + B X^{\text{seq}}_t,\\
     y_t = Ch_t + D X^{seq}_t .
\end{align}



\noindent \textbf{Dynamic Convolution.} 
Mamba primarily leverage manually designed scans to flatten image patches into one-dimensional sequences locally or globally. This approach disrupts the original semantic spatial adjacency of the image and lacks flexibility, making it difficult to capture complex image structures. 
To balance structure preservation and spatial semantic perception, we employ an input-dependent dynamic convolution to explore spatial adjacency representation, complementing the global modeling of reordering Mamba. 
Specifically, given an input feature map $X_2 \in \mathbb{R}^{H \times W \times C_2}$, we first apply global average pooling to extract channel-wise statistics. These are processed by two point-wise convolutions to compute group-wise attention weights, depicted as 
\begin{align}
\mathbf{A}' &= \text{Conv}_{1 \times 1}^{\frac{C_2}{r} \rightarrow (G \times C_2)} \left( \text{Conv}_{1 \times 1}^{C \rightarrow \frac{C_2}{r}} (\text{AvgPool}(X_2)) \right) , \\
\mathbf{A} &= \text{Softmax}(\text{Reshape}(\mathbf{A}'))  , \\
\mathbf{W} &= \sum_{i=1}^{G} \mathbf{P}_i \mathbf{A}_i ,
\end{align}
where $\mathbf{P}_i$ denotes the kernel bank for group $i$, and $\mathbf{W}$ represents the final dynamic convolution weights. 

\noindent \textbf{Fusion Unit} To integrate the complementary features from reordering Mamba and dynamic convolution, we concatenate them in the channel dimension and feed them into a lightweight channel fusion head, finally performing residual superposition.

\begin{align}
    F_{cat} = Contact(F_g,F_l),\\
    F_{out} = F_{fuse}(F_{cat}) + F_{cat},
\end{align}
where $F_g$ and $F_l \in \mathbb{R}^{H \times W \times \frac{C}{2}}$ refer to outputs from the Mamba and Conv branches. 
$F_{fuse}$ adopts a depth-separable 3×3 convolution to capture local spatial consistency, followed by channel mixing using 1×1 → GroupNorm → GELU → 1×1 → GroupNorm → GELU. GroupNorm is used between layers to adapt to small batch training and improve stability. This design integrates cross-channel information at extremely low computational cost.

\begin{figure}
    \centering
    \includegraphics[width=0.985\linewidth]{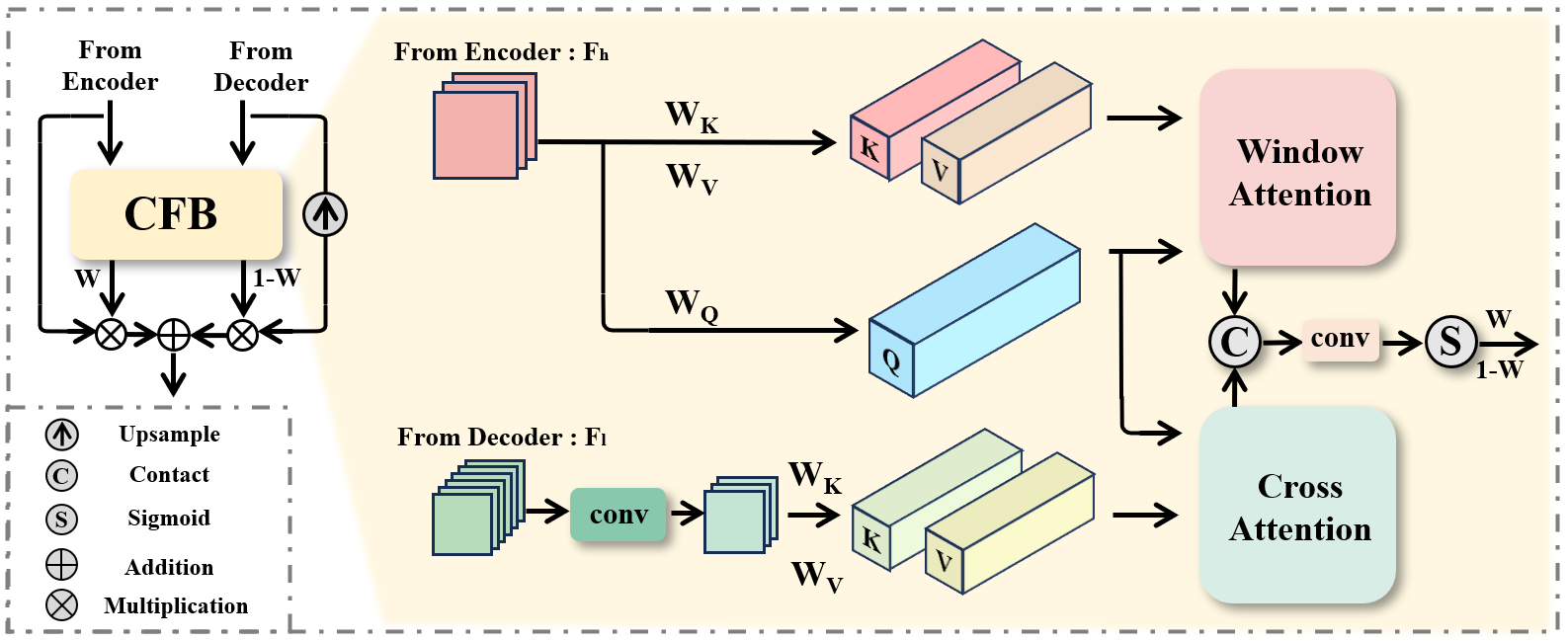}
    \caption{ Illustration of the proposed Cross-level Feature Bridging (CFB) module.}\vspace{-2mm}
    \label{fig:cfb}
\end{figure}

\subsection{Cross-Feature Bridge (CFB)}

To address the challenges of color distortion and loss of detail in underwater images, we propose a Cross-Feature Bridge (CFB) module, as illustrated in Figure ~\ref{fig:cfb}.
This module performs content-aware feature integration through a hybrid gating strategy that dynamically balances and fuses knowledge from the encoder and decoder. In this way, the structural details and the global context are fully integrated, which facilitates spatial fidelity and color correction.

\noindent \textbf{Local Detail Enhancement.}
To preserve local details and spatial fidelity of the encoder feature $F_h \in \mathbb{R}^{H \times W \times C}$, we employ local window self-attention to model fine-grained relationships within $F_h$:
\begin{equation}
\mathbf{F}'_h = \text{Softmax} \left( \frac{Q_h K_h^T}{\sqrt{d_k}} \right) V_h.
\end{equation}
This reinforces important structural patterns in high-resolution features.

\noindent \textbf{Global Context Enhancement.}
To further improve the robust representation of decoder, we apply cross-attention to explore the complementary global context information from the encoder feature. Specifically, 
we first align channels by mapping the decoder feature $F_l \in \mathbb{R}^{\frac{H}{2} \times \frac{W}{2} \times 2C}$ to $F_{l1} = Conv_{1\times1}(F_l)\in \mathbb{R}^{\frac{H}{2} \times \frac{W}{2} \times C}$, and then 
the encoder feature serves as the query and the decoder feature as key/value based on $F_{l1}$ to learn the cross-layer correlation, depicted as
\begin{equation}
\mathbf{F}'_{l1}  = \text{Softmax} \left( \frac{Q_h K_l^T}{\sqrt{d_k}} \right) V_l.
\end{equation}

\noindent \textbf{Spatial-Channel Gated Fusion.}
Instead of simply summing or concatenating the two attentional features, we introduce a gated fusion mechanism that learns adaptive spatial-channel weights to balance structural details and semantic context:
\begin{gather}
W = \sigma \left( \text{Conv}_{1\times1}\left( [\mathbf{F}'_h, \mathbf{F}'_{l1} ] \right) \right), \\
F{\text{out}} = W \odot F_h + (1 - W) \odot \text{Upsample}(F_l).
\end{gather}

Here, $W \in \mathbb{R}^{H \times W \times C}$ is a spatial-channel attention map that content-adaptively blends the two features at each spatial location and channel. This allows the model to correct color distortions by leveraging global context from the decoder while preserving local structural details from the encoder.

\setlength{\tabcolsep}{3.5pt}
\renewcommand{\arraystretch}{1.25}
\begin{table*}[!htb]
\centering
\caption{
Quantitative comparisons on three full reference test sets using PSNR/SSIM/MSE and UCIQE.
The top three results are marked with {\color{red} RED}, {\color{blue} BLUE}, and  {\color{green} GREEN}, respectively.}
\label{tab:full-ref}
\resizebox{\textwidth}{!}{
\begin{tabular}{l|cccc|cccc|cccc|c|ccc}
\toprule
\multicolumn{1}{c|}{Method} &
\multicolumn{4}{c|}{\cellcolor{blue!5}\textbf{UIEB}} &
\multicolumn{4}{c|}{\cellcolor{blue!5}\textbf{LSUI}} &
\multicolumn{4}{c|}{\cellcolor{blue!5}\textbf{EUVP}} &
\multicolumn{1}{c|}{\cellcolor{blue!5}\textbf{Average}} &
\multicolumn{3}{c}{\cellcolor{blue!5}\textbf{Complexity}} \\
\cmidrule(lr){2-5}\cmidrule(lr){6-9}\cmidrule(lr){10-13}\cmidrule(lr){14-14}\cmidrule(lr){15-17}
& PSNR$\uparrow$ & SSIM$\uparrow$ & MSE$\downarrow$ & UCIQE$\uparrow$
& PSNR$\uparrow$ & SSIM$\uparrow$ & MSE$\downarrow$ & UCIQE$\uparrow$
& PSNR$\uparrow$ & SSIM$\uparrow$ & MSE$\downarrow$ & UCIQE$\uparrow$
& PSNR$\uparrow$ & Params & GFLOPs & Times\\
\midrule
U-Trans~\cite{10129222}             & 20.77 & 0.801 & 92.08 & 0.576 & 24.23 & 0.842 & 80.45 & 0.574 & 25.33 & 0.823 & 74.21 & 0.579  & 23.44 & 65.60 & 66.20  & 0.102\\
Semi-UIR~\cite{Semi-UIE}            & 23.49 & 0.860 & 79.86 & \color{green} 0.612 & 27.46 & 0.883 & 65.31 & 0.590 & 27.85 & 0.876 & 60.57 & 0.586  & 26.27 & 3.31  & 72.88  & 1.280\\
UIE-DM~\cite{Tang2023UnderwaterIE}  & 20.90 & 0.817 & 85.41 & 0.593 & 25.49 & 0.853 & 66.01 & 0.582 & 26.86 & 0.859 & 57.04 & 0.591  & 24.42 & 10.70 & 66.89 & 0.321\\
GUPDM~\cite{GUPDM}                  & 21.84 & 0.844 & 87.54 & 0.584 & 25.38 & 0.864 & 74.53 & 0.573 & 25.21 & 0.844 & 74.46 & 0.577  & 24.14 & 1.49  & 95.80 & 1.134\\
CDF~\cite{CDF}                      & 22.64 & 0.809 & 81.81 & 0.609 & 22.71 & 0.814 & 80.16 & \color{red}0.597& 19.62 & 0.782 & 91.11 & {\color{red} 0.608}  & 21.66 & 15.90 & 57.52 & 0.070\\
CECF~\cite{CECF}                    & 21.59 & 0.815 & 87.16 & 0.609 & 23.79 & 0.839 & 79.50 & {\color{green} 0.592}& 23.36 & 0.813 & 79.41 &  \color{green}0.592  & 22.91 & 39.69 & 83.58 & 0.003\\
HCLR~\cite{HCLR}                    & 23.81 & 0.861 & 76.91 & 0.603 & 27.55 & 0.877 & 57.56 & 0.586 & 27.90 & 0.857 & 52.23 & 0.583  & 26.42 & 4.87  & 401.90 &0.088\\
wFormer~\cite{wFormer}              & 23.85 & {\color{blue} 0.878} & 76.30 & 0.609 & 27.52 & \color{green} 0.897 & 57.65 & 0.591 & 28.08 & 0.891 & 50.48 & 0.589  & 26.48 & 27.10 & 49.80 &0.158\\
UIR-PK~\cite{UIR-PK}                & 21.50 & 0.827 & 87.18 & 0.591 & 24.03 & 0.851 & 77.20 & 0.582 & 24.43 & 0.848 & 72.61 & 0.589  & 23.32 & 1.84  & 13.67 &0.006\\
wMamba~\cite{guan2024watermamba}    & 23.88 & 0.865 & 76.79 &  0.610 &  \color{green} 28.33 & 0.889 & {\color{green}51.78} & 0.588 & {\color{green} 28.76} & 0.881 &  {\color{green}44.56} & 0.587  & \color{green}26.99 & 3.69  & 7.53 &0.045\\
MambaIR~\cite{MambaIR}              & {\color{green} 23.94} & {\color{blue} 0.878} & {\color{green}75.89} & 0.611 & {\color{green} 28.33} & \color{blue}0.901 &  53.16 & 0.591 & 28.65 & {\color{green} 0.893} &  46.76 & 0.590  & 26.97 & 25.92 & 137.57 &0.168\\
\midrule
\rowcolor{gray!10}
\textbf{VRS-UIE-S (Ours)} & \color{blue} 24.34 & \color{green} 0.875 & \color{blue} 73.62 & \color{blue} 0.615 & \color{blue} 28.77 & \color{blue} 0.901 & \color{blue} 49.80 & \color{green} 0.592 & \color{blue} 29.56 & \color{blue} 0.899 & \color{blue}41.30 & 0.591 & \color{blue} 27.55 & 2.06 & 2.55 & 0.058\\
\rowcolor{gray!10}
\textbf{VRS-UIE (Ours)} & {\color{red}24.91} & {\color{red}0.879} & {\color{red}70.55} & {\color{red}0.616} & {\color{red}29.00} & {\color{red}0.904} &  {\color{red}48.47} & {\color{blue}0.594} & {\color{red}29.74} & {\color{red}0.902} &  {\color{red}40.28} & {\color{blue}0.595} & \color{red}27.88 &  7.02 & 36.47  & 0.153\\
\bottomrule
\end{tabular}}
\end{table*}

\subsection{Loss Functions and Constraints}
Since underwater images typically exhibit both structural degradation and texture blur, we design a composite loss function that jointly optimizes pixel-wise fidelity, perceptual consistency, and edge preservation. The total loss consists of four components: $\ell_1$ reconstruction loss, SSIM-based structure loss,  edge-aware loss, and multi-granularity KL loss (mentioned in MVGL, denote as $L_{mvgl}$).

\noindent \textbf{Pixel-Wise Loss.}
To ensure content-level consistency and suppress over-enhancement artifacts, we adopt the standard $\ell_1$ loss:
\begin{equation}
L_{\text{L1}} = \frac{1}{N} \sum_{i=1}^N \left| \hat{y}_i - y_i \right|_1.
\end{equation}

\noindent \textbf{Structural Similarity Loss.}
We incorporate SSIM loss\cite{ssimloss} to enhance the visual similarity in terms of luminance, contrast, and structural layout:
\begin{equation}
L_{\text{SSIM}} = 1 - \text{SSIM}(\hat{y}_i, y_i).
\end{equation}

\noindent \textbf{Edge Preservation Loss.}
To preserve high-frequency details and suppress structural blur, we use an edge-aware loss computed from Laplacian-based edge maps:
\begin{equation}
L_{\text{edge}} = \sqrt{ \left( \text{Edge}(\hat{y}_i) - \text{Edge}(y_i) \right)^2 + \epsilon^2 },
\end{equation}
where $\text{Edge}(\cdot)$ denotes a fixed Laplacian edge extractor \cite{charbonnier}.

\noindent\textbf{Total Objective Loss.}
The final loss is a weighted sum:
\begin{equation}
L_{\text{total}} = \lambda_1 L_{\text{L1}} + \lambda_2 L_{\text{SSIM}} + \lambda_3 L_{\text{edge}} + \lambda_4 L_{\text{mvgl}},
\end{equation}
where we empirically set $\lambda_1 = 8$, $\lambda_2 = 1$, $\lambda_3 = 4$, and $\lambda_4 = 2e^{-3}$ to balance fidelity and structural preservation.

\section{Experiments}
\subsection{Implementation Details}

Our experiments are conducted on two widely used real-world underwater image datasets: UIEB~\cite{C60} and LSUI~\cite{10129222}. 
The UIEB dataset consists of 890 paired underwater images, where the reference image for each scene is selected through a human voting process involving 50 volunteers from outputs generated by 12 UIE algorithms. 
The LSUI dataset further expands the scene diversity with 4,274 underwater images captured under varying lighting conditions, water types, and object categories. 
Its reference images are selected from 18 algorithm outputs and finalized through a ranking by 20 annotators.

For training, we utilize 830 randomly sampled image pairs from UIEB and 3,300 from LSUI. All images are resized to a resolution of 256$\times$256. 
Our model is implemented in PyTorch 2.4.1 and trained on a single NVIDIA RTX 3090 GPU using the AdamW optimizer with an initial learning rate of $1 \times 10^{-4}$, and momentum parameters $(\beta_1, \beta_2) = (0.9, 0.999)$. The network is trained for 300,000 iterations with a batch size of 2, and a cosine annealing scheduler is employed for learning rate decay.

\setlength{\tabcolsep}{3.5pt}
\renewcommand{\arraystretch}{1.25}
\begin{table*}[!htb]
\centering
\caption{Quantitative comparisons on three no reference test sets:U45/SQUID-16 and UCCS using UCIQE, UIQM, URanker and CLIP-IQA. The top three results are marked with {\color{red} RED}, {\color{blue} BLUE}, and  {\color{green} GREEN}, respectively.}
\label{tab:no-ref}
\resizebox{\textwidth}{!}{
\begin{tabular}{l|cccc|cccc|cccc}
\toprule
\multicolumn{1}{c|}{Method} &
\multicolumn{4}{c|}{\cellcolor{blue!5}{\textbf{U45}}} &
\multicolumn{4}{c|}{\cellcolor{blue!5}{\textbf{SQUID-16}}} &
\multicolumn{4}{c}{\cellcolor{blue!5}{\textbf{UCCS}}}\\
\cmidrule(lr){2-5}\cmidrule(lr){6-9}\cmidrule(lr){10-13}
& UCIQE$\uparrow$ & UIQM$\uparrow$ & URanker$\uparrow$ & CLIP-IQA$\uparrow$
& UCIQE$\uparrow$ & UIQM$\uparrow$ & URanker$\uparrow$ & CLIP-IQA$\uparrow$
& UCIQE$\uparrow$ & UIQM$\uparrow$ & URanker$\uparrow$ & CLIP-IQA$\uparrow$ \\
\midrule
U-Trans~\cite{10129222}            & 0.567 & 3.105 & 1.242 & 0.435 & 0.544 & 2.230 & 0.554 & 0.524 & 0.547 & 3.040 & 1.331 & 0.380 \\
Semi-UIR\cite{Semi-UIE}            & \color{green}0.601 & 3.131 & \color{red}2.031 & 0.510 & \color{green}0.588 & 2.309 & \color{blue}1.110 & \color{red}0.586 & \color{green}0.571 & 2.947 & \color{red}1.723 & 0.393 \\
UIE-DM~\cite{Tang2023UnderwaterIE} & \color{blue}0.603 & 3.108 & 1.810 & \color{green}0.533 & \color{red}0.659 & 0.519 & -0.070 & 0.560 & \color{red}0.628 & 0.538 & 0.642 & \color{red}0.504 \\
GUPDM\cite{GUPDM}                  & 0.573 & 3.028 & 1.591 & 0.528 & 0.577 & \color{green}2.486 & 0.952 & \color{green}0.568 & 0.552 & 2.981 & 1.415 & 0.408 \\
CDF\cite{CDF}                      & 0.589 & 2.972 & 1.690 & 0.443 & 0.575 & 2.265 & 0.918 & 0.337 & 0.561 & 2.800 & 1.347 & 0.279 \\
CECF\cite{CECF}                    & 0.592 & 3.101 & \color{green}1.872 & 0.507 & 0.578 & 2.476 & 0.992 & 0.432 & 0.567 & \color{red}3.013 & 1.607 & 0.346 \\
HCLR\cite{HCLR}                    & 0.592 & \color{red}3.168 & 1.729 & \color{red}0.554 & 0.576 & 2.449 & 0.947 & \color{blue}0.577 & 0.565 & \color{blue}2.993 & 1.503 & 0.389 \\
WFormer\cite{wFormer} &0.592 &3.084 &1.830 &0.517  &0.577 &2.373 &\color{green}1.102 &0.516  &0.564 &2.968 &\color{green}1.630 &0.345\\
UIR-PK~\cite{UIR-PK}               & 0.580 & 2.964 & 1.170 & 0.472 & 0.574 & 2.406 &1.055 & 0.506 & 0.549 & 2.861 & 1.248 & 0.374 \\
WMamba  \cite{guan2024watermamba}  & 0.596 & 3.099 & 1.788 & 0.477 & 0.580 & 2.445 & 1.034 & 0.491 & 0.567 & 2.974 & 1.519 & 0.327 \\
MambaIR \cite{MambaIR}             & 0.599 & \color{green}3.149 & 1.800 & 0.486 & 0.577 & \color{red}2.504 & 0.911 & 0.512 & 0.552 & 3.001 & 1.117 & \color{blue}0.432 \\
\midrule
\rowcolor{gray!10}
VRS-UIE-S (Ours) &\color{green}0.601 &3.102 & 1.828 & 0.490 & \color{green}0.588 &2.475 & 1.063 & 0.498 & \color{green}0.571 & 2.950 & 1.557 & 0.396 \\
\rowcolor{gray!10}
VRS-UIE (Ours) & \color{red}0.604 & \color{blue}3.154 & \color{blue}1.927 & \color{blue}0.536 & \color{blue}0.589 & \color{blue}2.489 & \color{red}1.134 & 0.541 & \color{blue}0.572 & \color{green}2.984 & \color{blue}1.647 & \color{green}0.419 \\
\bottomrule
\end{tabular}}
\end{table*}

\begin{figure*}[!htb]
    \centering
    \includegraphics[width=1\linewidth]{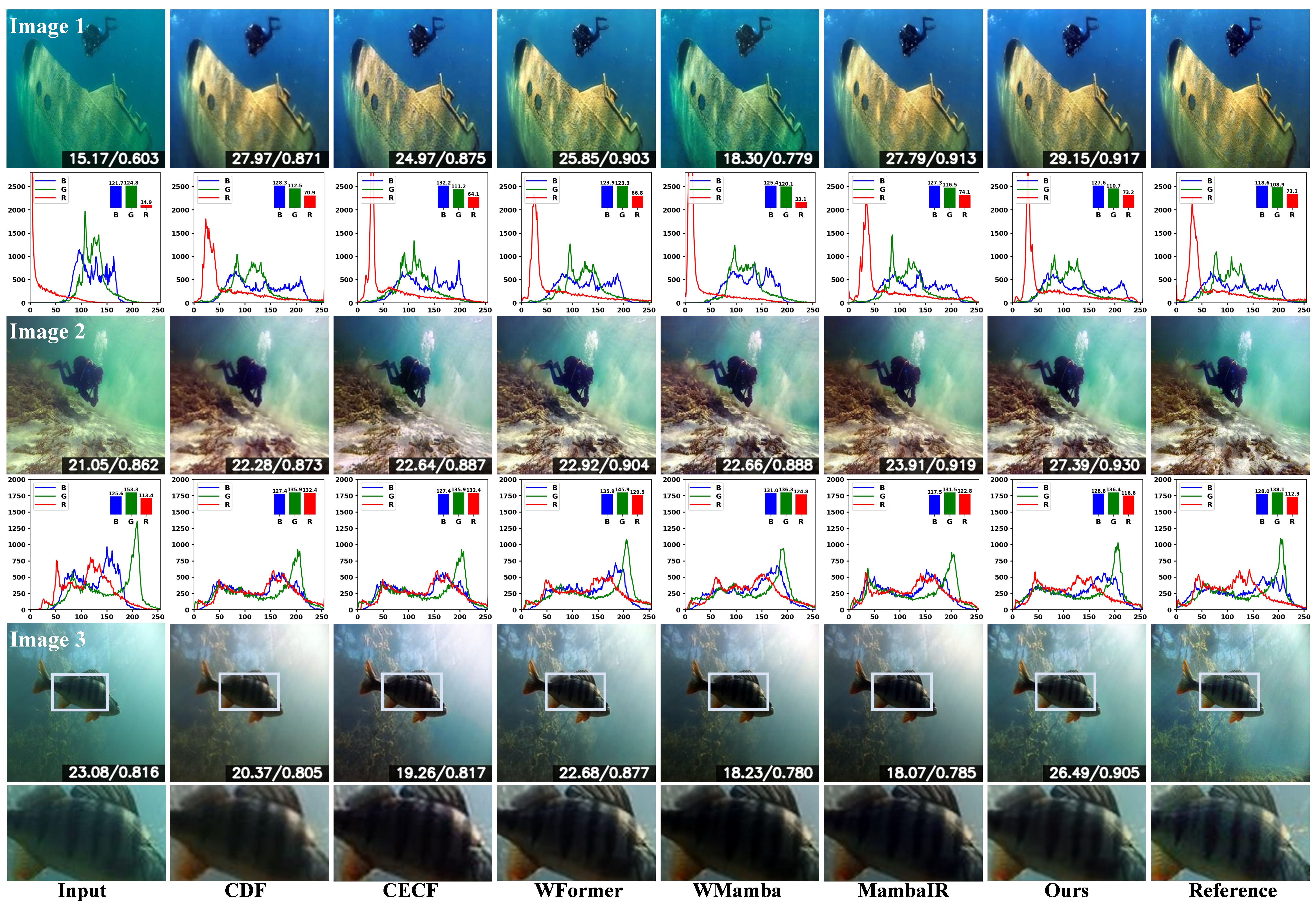}
    \caption{Full reference comparison on different scenes. For each scene we report PSNR/SSIM (higher is better) with respect to the reference shown in the last column. The curve plot displays an image histogram, with a bar graph in the upper-right corner showing the average pixel value for each channel.
    }\vspace{-4mm}
    \label{fig:ps}
\end{figure*}

\begin{figure*}[!htb]
      \centering
    \includegraphics[width=1\linewidth]{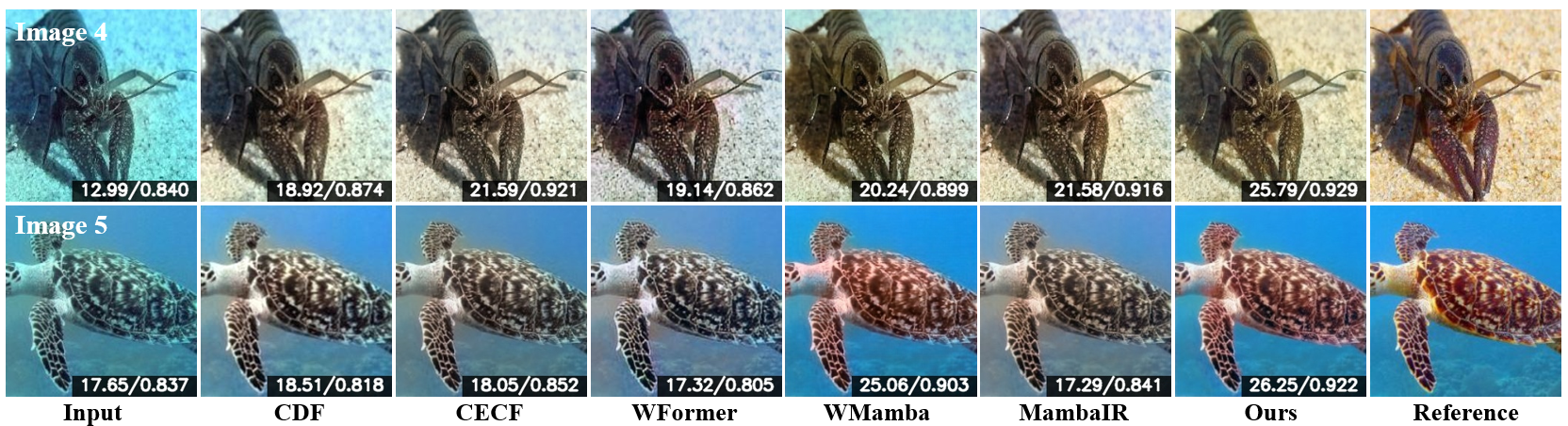}
    \caption{Restoration Under Extreme Degradation. For each scene we report PSNR/SSIM (higher is better) with respect to the reference shown in the last column. }
    \label{fig:limit}
\end{figure*}

\begin{figure*}[!htb]
    \centering
    \includegraphics[width=1\linewidth]{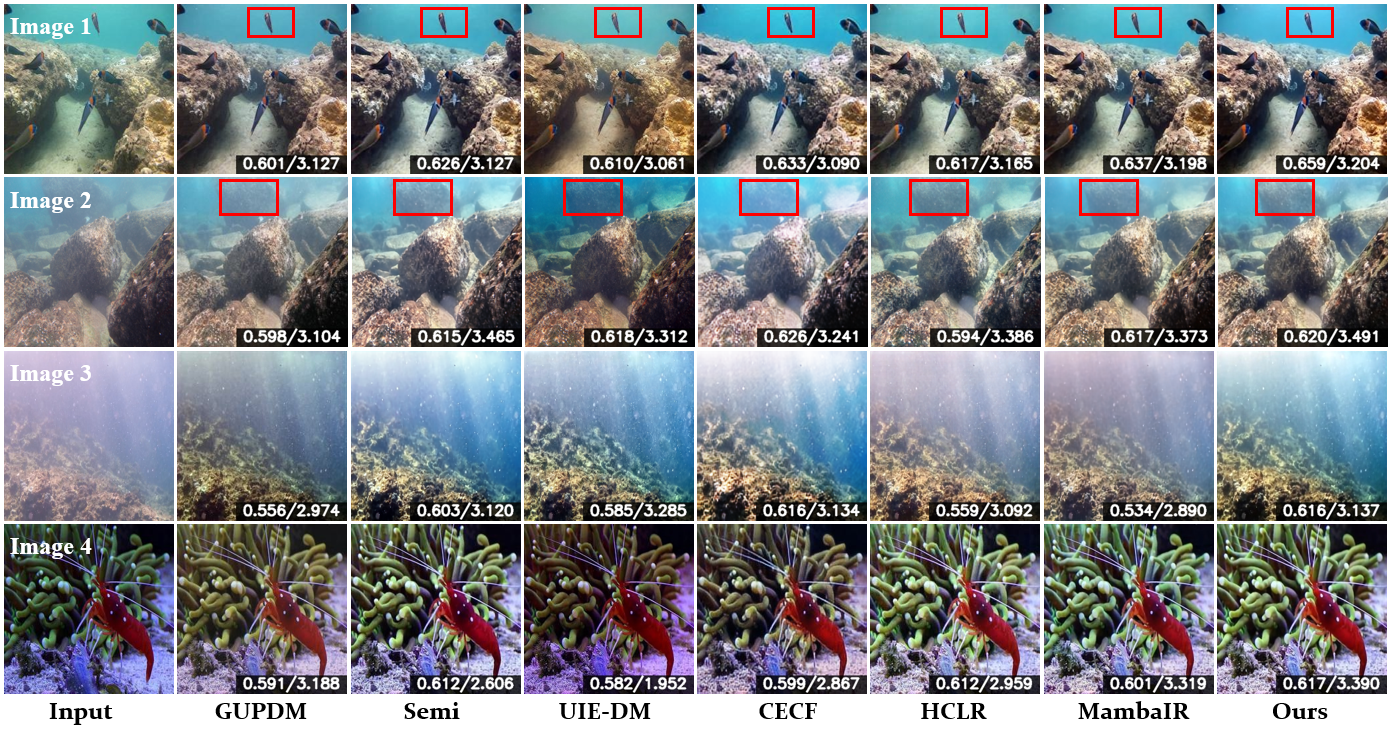}
    \caption{No Reference comparison on different scenes. For each scene we report UCIQE/UIQM (higher is better). 
    }\vspace{-4mm}
    \label{fig:2u}
\end{figure*}

\subsection{Experimental Settings}

\noindent \textbf{Benchmarks.} We evaluate our method on both full-reference (FR) and non-reference (NR) underwater enhancement benchmarks.
(1) \textit{FR test sets.} We use 60, 975, and 481 paired images from the UIEB, LSUI, and EUVP~\cite{EUVP} datasets, respectively. These datasets provide ground-truth references to assess pixel-level fidelity and structural similarity. 
(2) \textit{NR test sets.} To evaluate generalization in diverse underwater scenarios, we test on three real-world unpaired benchmarks: 45 images from the U45~\cite{U45}, 16 images from the SQUID~\cite{SQUID}, and 330 images from the UCCS~\cite{UCCS} datasets, which are used to assess perceptual enhancement quality.

\noindent \textbf{Evaluation Metrics.} 
We adopt Peak Signal-to-Noise Ratio (PSNR), Structural Similarity Index (SSIM), and Mean Squared Error (MSE) for full-reference comparison, while the UCIQE~\cite{uciqe}, UIQM~\cite{uiqm}, Uranker~\cite{Uranker} and CLip-IQA ~\cite{clip-iqa} are used as non-reference metrics. 
These metrics jointly assess underwater perceptual quality from contrast, color fidelity, naturalness, and sharpness perspectives.

\noindent \textbf{Compared Methods.} We compare our proposed VRS-UIE against eleven state-of-the-art underwater image enhancement (UIE) approaches. These methods involve U-Trans~\cite{10129222}, Semi-UIR~\cite{Semi-UIE}, UIE-DM~\cite{Tang2023UnderwaterIE}, GUPDM~\cite{GUPDM}, CDF~\cite{CDF}, CECF~\cite{CECF}, HCLR~\cite{HCLR}, WFormer~\cite{wFormer}, UIR-PK~\cite{UIR-PK}, WMamba~\cite{guan2024watermamba} and MambaIR~\cite{MambaIR}.

\subsection{Quantitative Comparisons}
As shown in Table \ref{tab:full-ref}, our proposed method consistently outperforms all previous approaches in three FR benchmarks on UIEB, LSUI, and EUVP datasets.
Our high-accuracy model (VRS-UIE) achieves the highest PSNR and SSIM on all three sets and the lowest Mean Squared Error (MSE), surpassing the SOTA method Wmamba~~\cite{guan2024watermamba}  by 0.89~dB on average. In terms of perceptual fidelity, as measured by UCIQE, our model achieves the highest scores on UIEB (0.616) and the second best score on LSUI (0.594) and EUVP (0.595), slightly behind CDF~\cite{CDF} (0.597 / 0.608). 

Based on VRS-UIE, to offer a favorable efficiency trade-off, we develop a compact and high-efficiency variant (VRS-UIE-S). 
Specifically, we adjust the channels (32 to 24) and blocks per stage, and introduce wavelet transform to project the feature extraction in the low-resolution frequency space to improve inference speed. 
With only 2.06~M parameters and 2.55 GFLOPs per image with resolution $256\times 256$, our VRS-UIE-S runs in 0.058 s on a single RTX 3090. Benefiting from the elaborate Value-Driven Reordering Scanning, Mamba-Conv Mixer and Cross-Feature Bridge,  VRS-UIE-S achieves impressive enhancement results, surpassing the light-weight UIE model (wMamba~\cite{guan2024watermamba}) by 0.56~dB in PSNR on average. It is noted that wMamba significantly requires addition 79\% and 195\% of the model parameters and computational cost in comparison to our VRS-UIE-S method. 
The high efficiency of VRS-UIE-S makes it practical for real applications.

We further evaluate all methods on three NR datasets with four metrics (UCIQE, UIQM, URanker and CLIP-IQA). As shown in Table \ref{tab:no-ref}, our VRS-UIE shows impressive performance across all metrics on the U45 dataset, gaining the highest UCIQE (0.604), UIQM (3.154), and URanker (1.927) scores, and a competitive CLIP-IQA score (0.536). In the SQUID-16 dataset, our method achieves the top URanker score (1.134), while delivering competitive scores for UCIQE (0.589) and UIQM (2.489). In the UCCS dataset, our method obtains the competitive scores, with the second best UCIQE (0.572) and URanker (1.647). These results from different metric emphases demonstrate the robustness and generalization of our model under diverse scenes.


\subsection{Qualitative Comparisons}

To further illustrate the effectiveness of our approach, we present the visual comparison on representative underwater scenes. The first panel showcases examples selected based on FR metrics, and the second panel showcases examples selected based on NR metrics. 

In the FR-selected panel (Figure \ref{fig:ps} and Figure \ref{fig:limit}), 
especially for blue/green-shifted underwater scenes (the 1-2$^{th}$ scenarios), our method achieves the best PSNR/SSIM. 
The statistical histograms of RGB channels further show closer agreement with the reference—peaks occur at similar intensities and with better distribution consistence, while blue/green spikes are suppressed. 
In the backlight case (the 3$^{th}$ scenario), our result lifts exposure and preserves stripe textures on the fish, while other methods, such as CECF, HCLR, and WMamba, exhibit under-enhancement or loss of textural detail.
We further analyze the recovery effect under extreme degenerate conditions in Figure \ref{fig:limit}, our method produces a spatially coherent, warm color tone that is closer to the reference image, whereas others correct only partially and exhibit spatial inconsistency in the 4$^{th}$ scenario. In the 5$^{th}$ scenario, our method better recovers the natural coloration of the turtle. By contrast, CDF/CECF can only partially restore color, and WMamba preserves the global tone but loses details on the distant fore-flipper.

In the NR-selected panel (Figure \ref{fig:2u}), these tendencies remain visible in a different set of strong NR performers.  
Our outputs effectively restore the vivid colors of the fish while better resolving color casts and blurring at a distance (the 1$^{th}$ scenario). In addition, our method suppresses back-scatter effectively without introducing a reddish bias (the 2$^{th}$ and 3$^{th}$ scenarios). 
It also delivers balanced illumination in low-light conditions while providing more saturated yet natural color reproduction for various subjects such as shrimp and aquatic plants (the 4$^{th}$ scenario).

Overall, these visual results demonstrate that our approach provides superior color correction and illumination adjustment while effectively preserving fine structural details, which complements the quantitative FR and NR results.

\subsection{Analysis on MVGL}
To comprehensively evaluate the proposed Multi-Granularity Value Guidance Learning (MVGL) 
module, we perform comprehensive analyzes from three perspectives: visualization of DINO $L_2$-norm features, a comparative analysis of alignment mechanisms, and a sensitivity analysis of the hyperparameter $k$ governing the deformable sampling kernel size.

\begin{figure}
    \centering
    \includegraphics[width=0.985\linewidth]{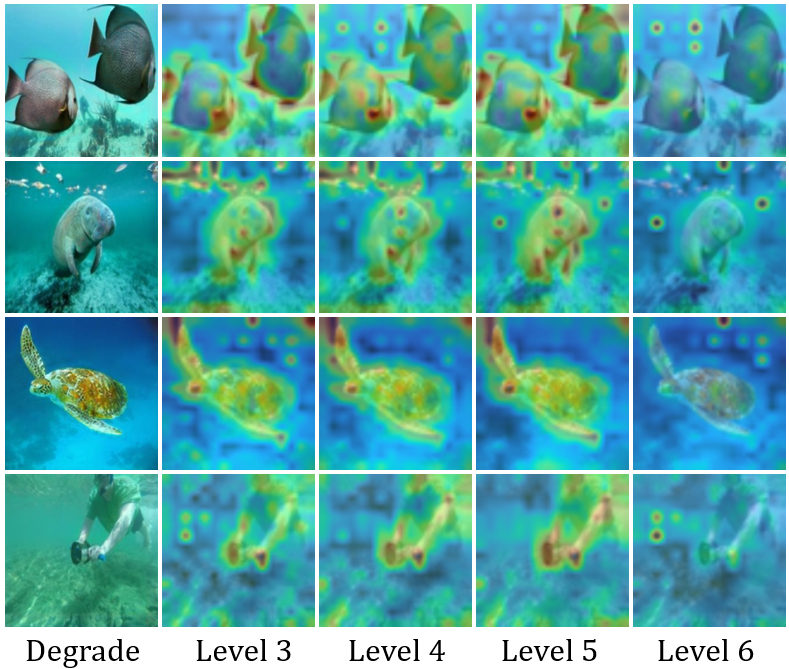}
    \caption{Visual comparisons results on Stage-wise DINO L2 norm feat. At the appropriate level, dino feat can identify areas with sparse and more valuable prospects underwater.}
    \vspace{-2mm}
    \label{fig:dino}
\end{figure}

\noindent \textbf{Stage-wise DINO Evidence.}
Figure \ref{fig:dino} visualizes the $L_2$-norm of DINO features from Levels 3–6 overlaid at a resolution of $64\times64$. Across diverse underwater scenes, high-response regions are consistently localized to object boundaries, salient foreground objects, and chromatic anchors. 
Shallow levels exhibit a bias towards edges, whereas L5 demonstrates greater region-level consistency. 
This behavior aligns with the typical roles of U-Net stages: high-resolution stages recover fine details and textural elements, while low-resolution stages enforce semantic consistency. Consequently, we adopt a multi-granularity supervision strategy: high-resolution encoder/decoder stages are supervised by L3/L4 features, and low-resolution/bottleneck stages by L5. Since L6 exhibits unstable peaks in underwater scenes, it is therefore excluded from providing supervision signals for MVGL.


\noindent \textbf{Comparative Analysis of Guidance Mechanisms.}
We study four ways to transfer a token-level value prior \(D\) to a dense value map \(V\) for scan reordering.
Let \(D\in\mathbb{R}^{h\times w}\) with \(h=w=16\) be the token prior computed from DINOv2 features by taking the channel-wise \(L_2\) norm, and let \(S\in\mathbb{R}^{H\times W}\) be the implicit sampling frequency map.
We compare:
(a) lifting \(D\) to \(H\times W\) and ranking pixels by the lifted prior \(D^\uparrow\);
(b) average-pooling \(S\) to the token lattice $(16 \times 16)$ and minimizing \(\mathrm{KL}(D\,\|\,\mathrm{pool}( S))\);
(c) lifting \(D\) to \(H\times W\) and minimizing \(\mathrm{KL}(D^\uparrow\,\|\, S)\) on the image grid;
(d) ours (multi-coverage masked KL guidance), which applies KL within top-k masks at several coverage levels to encourage focus on valuable regions and adapt distribution variability.

\begin{figure}
    \centering
    \includegraphics[width=0.985\linewidth]{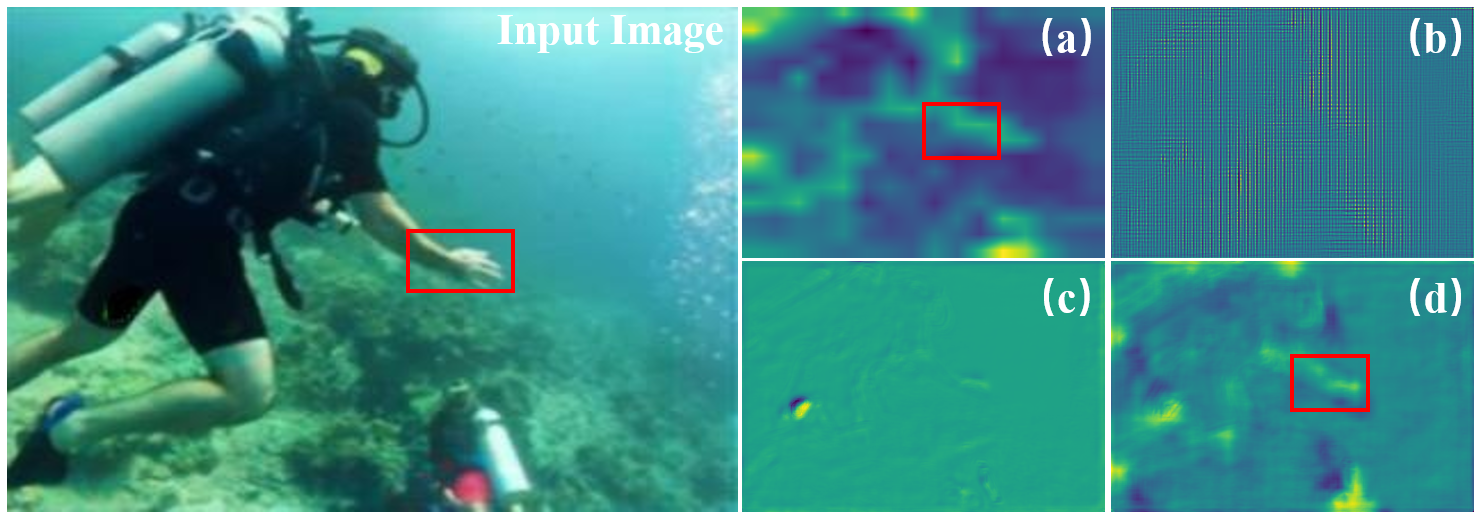}
    \caption{Value maps obtained with different guidance mechanisms (a–d); the brighter pixel indicates the higher value region.}
    \label{fig:learn}
\end{figure}

Figure \ref{fig:learn} shows the resulting value map used for reordering in the SSM.
Mechanism (a) exhibits geometric misalignment, as interpolation artifacts token responses across neighboring pixels (\emph{e.g.}, the diver's hand is visibly displaced, as indicated by the red box).
Mechanism (b) ties supervision to the token lattice, producing checkerboard-like priors that lack pixel-accurate localization. 
Mechanism (c) treats all pixels equally within the $H \times W$ distribution, preventing the value map from learning corresponding high-value regions at high resolutions.
In contrast, our Mechanism (d) - MVGL adopts multi-granularity kl divergence to focus on valuable contents. When combined with deformable offset sampling, sharp and geometrically aligned peaks are recovered along edges and foreground regions (the structure within the red box is crisp and correctly aligned). 

Quantitatively (Table \ref{tab:Alignment_strategies}), methods (a)/(b)/(c) achieve 28.56/28.01/28.90 dB PSNR and 0.898/0.895/0.898 SSIM, whereas MVGL reaches 29.00 dB and 0.904. This corresponds to +0.44 dB PSNR over direct interpolation with a SSIM increase from 0.898 to 0.904.

\begin{table*}[ht]
\centering
\caption{Comparison of Different Configurations in Our Architecture.}
\begingroup
\setlength{\tabcolsep}{6pt}
\renewcommand{\arraystretch}{1.2}
\begin{tabular}{l|ccc|cccccc}
\toprule
\rowcolor{blue!10}
\textbf{Model Variant} & \textbf{MVGL} & \textbf{Dynamic Conv} & \textbf{CFB}
& \textbf{PSNR}$\uparrow$ & \textbf{SSIM}$\uparrow$ & \textbf{UCIQE}$\uparrow$ & \textbf{UIQM}$\uparrow$& \textbf{Params} & \textbf{FLOPs} \\
\midrule
(a) only reordering Mamba & \checkmark & $\times$ & $\times$ & \textbf{29.06} & 0.904 & 0.592 & 2.887 & 13.48 & 60.69 \\
(b) only dynamic convolution & $\times$ & \checkmark & $\times$ & 28.27 & 0.899  & 0.591 & 2.888 & 4.93 & 24.66 \\
(c) only Mixer & \checkmark & \checkmark & $\times$ & 28.79 & 0.902 & 0.593 & 2.883 & 6.90 & 35.76 \\
(d) dynamic convolution $\rightarrow$ 3$\times$3 Conv & \checkmark & $\times$ & \checkmark & 28.61 & 0.901  & 0.592 & 2.890 & 8.56 & 43.08 \\
\midrule
(e) MVGL $\rightarrow$ normal four-way scan & $\times$ & \checkmark & \checkmark & 28.77 & 0.903 & 0.593  & 2.883 &  7.29 &  31.59 \\
(f) reverse reordering for scan path & \checkmark & \checkmark & \checkmark & 28.88  & 0.903 & 0.593 & 2.879 & 7.02 & 36.47  \\
\midrule
\rowcolor{gray!10}
(g) \textbf{VRS-UIE} 
& \checkmark & \checkmark & \checkmark & 29.00  & \textbf{0.904} & \textbf{0.594} & \textbf{2.891} & 7.02 & 36.47  \\
\bottomrule
\end{tabular}
\endgroup
\label{tab:ablation_structure}
\end{table*}


\sisetup{
  table-number-alignment = center,
  detect-weight = true,          
  detect-inline-family = math,   
}

\begin{table}[t]
  \caption{Guidance strategies for injecting DINO cues (higher is better). $\Delta$ is the difference compared to (a).}
  \label{tab:Alignment_strategies}
  \centering
  \small
  \setlength{\tabcolsep}{2.5pt}
  \renewcommand{\arraystretch}{1.12}
  \begin{tabularx}{\columnwidth}{
    >{\raggedright\arraybackslash}X
    S[table-format=2.2]
    S[table-format=+1.3]
    S[table-format=+1.3]
    S[table-format=+1.2]
  }
    \toprule
    \rowcolor{blue!10}
    {Model Variant} &
    \multicolumn{1}{c}{PSNR (dB)$\uparrow$} &
    \multicolumn{1}{c}{$\Delta$} &
    \multicolumn{1}{c}{SSIM$\uparrow$} &
    \multicolumn{1}{c}{$\Delta$} \\
    \midrule
    Mechanism (a) & 28.56 &  0.00  & 0.898 &  0.00 \\
    Mechanism (b) & 28.01 & -0.55  & 0.895 & -0.03 \\
    Mechanism (c) & 28.90 & +0.34  & 0.898 & +0.00 \\
    \rowcolor{gray!10}
    \textbf{VRS-UIE} (d) & \textbf{29.00} & + 0.44 &  \textbf{0.904} & + 0.06 \\
    \bottomrule
  \end{tabularx}
\end{table}

\begin{table}[t]
\centering
\caption{Effects of the sampling hyperparameter $k$ for deformable alignment. $k^2$ is the number of sampling points per pixel. Best in bold.}
\begingroup
\setlength{\tabcolsep}{5pt}        
\small                        
\begin{tabular}{lcccc}
\toprule
\rowcolor{blue!10}
\textbf{Model Variant} & \textbf{PSNR}$\uparrow$ & \textbf{SSIM}$\uparrow$ & \textbf{Params(M)} & \textbf{FLOPs(G)} \\
\midrule
\rowcolor{gray!10}
$k=1$ & \textbf{29.00}  & \textbf{0.904} & 7.02 &  36.47 \\
$k=2$ & 28.79  & 0.901  & 7.03 & 36.61  \\
$k=3$ & 28.69 & 0.902   & 7.06 &  36.84 \\
\bottomrule
\end{tabular}
\endgroup
\label{tab:sampling_ablation}
\end{table}

\noindent \textbf{Effects of Sampling Hyperparameter $k$.}
We analyze the hyperparameter $k$, which defines the side length of the deformable sampling kernel and dictates the number of samples per pixel ($k^2$). 

As shown in Table \ref{tab:sampling_ablation}, performance degrades monotonically as \(k\) increases.
We hypothesize that a larger \(k\) disperses selections over multiple equivalent locations, flattening the sampling frequency map and weakening the value-ranking contrast, so valuable foreground regions become less emphasized.
We therefore adopt \(k{=}1\) by default.

\begin{figure}
    \centering
    \includegraphics[width=0.985\linewidth]{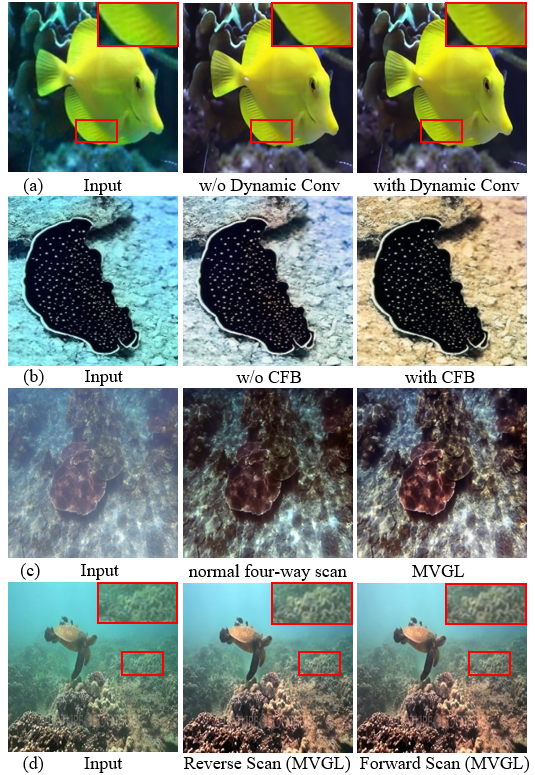}
    \caption{Visualization of a part of ablations. The red box in the upper right corner indicates the enlarged view of the corresponding area.}
    \label{fig:ab}
\end{figure}

\subsection{Ablation Study}

\noindent \textbf{Effects of Architectural Components.}
We evaluate the contribution of each module via controlled ablations that remove or modify one component at a time (Table \ref{tab:ablation_structure}; qualitative results in Figure \ref{fig:ab}).

\emph{Mixer.}
Variant (a) retains only the Reorder-Mamba path and variant (b) retains only the dynamic-convolution path.
The Mamba-only model achieves competitive scores but incurs higher parameters and FLOPs. Moreover, removing the dynamic branch leads to blurrier high-frequency details such as fish fins (Figure \ref{fig:ab}(a)).
The dynamic-only variant performs worse overall, indicating that local filtering alone is insufficient for long-range color and structure reasoning.

\emph{Cross-Feature Bridge (CFB).}
Replacing CFB with simple addition at the skip connections (variant (c)) degrades robustness under severe color cast. The naive addition in skip connection propagates artifacts into the decoder and weakens recovery quality (Figure \ref{fig:ab}(b)).

\emph{Dynamicity.}
Substituting the input-dependent convolution with a static $3{\times}3$ kernel (variant (d)) causes a clear drop in accuracy and perceptual fidelity, highlighting the benefit of sample-adaptive responses.

Taken together, these results show that value-driven global scanning, sample-adaptive local refinement, and cross-scale feature fusion are complementary, and their combination is critical to the performance–efficiency balance of VDS-UIE.



\noindent \textbf{Effects on Scan Method.} We conduct two experiments to evaluate the effectiveness of our proposed scan strategy. First, we replace our directional scan with a traditional four directional sequential scan in model variant (e), which leads to a performance drop. 
As shown in Figure \ref{fig:ab} (c), we observe that our method delivers clearer brightness and more effective color correction in underwater haze conditions.

Second, we maintain the MVGL to build the scan path but reverse the information flow. Interestingly, this modification also leads to a drop in performance. As shown in Figure \ref{fig:ab} (d), we observe that when the scanning sequence is reversed, distant areas appear more blurred compared to the forward sequence restoration results, with lower color bias correction.


\noindent \textbf{Effects of Loss Components.}
We ablate each term in the loss (Table\ref{tab:loss_ablation}) in order (a)–(e).
(a) Removal of $L_1$ produces the highest degradation, from 29.00/0.904 to 28.53/0.892, indicating that the pixel-wise fidelity is the primary anchor.
(b) Without $L_{\text{ssim}}$, performance drops to 28.83/0.899, showing that structural guidance is complementary to $L_1$.
(c) Removal of $L_{\text{edge}}$ yields 28.87/0.902, a small decrease in PSNR with approximate SSIM, suggesting that the edge term mainly sharpens the boundaries that global metrics capture weakly.
(d) Without $L_{\text{mvgl}}$, the multi-granularity KL loss provides explicit guidance and leaves only the implicit sampling path, reduces performance to 28.69/0.899. Reintroducing$L_{\text{mvgl}}$ yields gains of +0.31 dB and +0.005 SSIM, confirming that explicit guidance stabilizes value-guided ordering.
(d) Without $L_{\text{mvgl}}$, which designed to inject the explicit value prior, training relies only on the implicit branch and performance drops to 28.69/0.899. Reintroducing $L_{\text{mvgl}}$ yields +0.31 dB and +0.005 SSIM, showing that explicit guidance stabilizes value-driven ordering.
(e) The full model (VRS-UIE) achieves 29.00/0.904, confirming that all terms contribute, with $L_1$ providing the strongest supervision and $L_{\text{mvgl}}$ being the key to robust value-driven scanning.

\begin{table}[t]
\centering
\caption{Ablation on different loss components. ``\checkmark'' means the loss is used.}
\label{tab:loss_ablation}
\small
\setlength{\tabcolsep}{6pt}
\renewcommand{\arraystretch}{1.2}
\begin{tabular}{lcccccc}
\toprule
\rowcolor{blue!10}
Model Variant & $L_{1}$ & $L_{\text{ssim}}$ & $L_{\text{edge}}$ & $L_{\text{mvgl}}$ & PSNR & SSIM \\
\midrule
(a) w/o $L_{1}$ loss    & $\times$ & \checkmark & \checkmark & \checkmark & 28.53 & 0.892 \\
(b) w/o $L_{\text{ssim}}$ & \checkmark & $\times$ & \checkmark & \checkmark & 28.83 & 0.899 \\
(c) w/o $L_{\text{edge}}$ & \checkmark & \checkmark & $\times$ & \checkmark & 28.87 & 0.902 \\
(d) w/o $L_{\text{mvgl}}$ & \checkmark & \checkmark & \checkmark & $\times$  &  28.69 & 0.899 \\
\midrule
\rowcolor{gray!10}
(e) VRS-UIE   & \checkmark & \checkmark & \checkmark & \checkmark & \textbf{29.00} & \textbf{0.904} \\
\bottomrule
\end{tabular}
\end{table}

\section{Conclusion}
This paper presents a novel Value-Driven Reordering Scanning (VRS-UIE) framework for underwater image enhancement, which redefines conventional fixed-order scanning mechanisms by prioritizing content based on its semantic and structural significance, thereby facilitating more effective color inference and structural reconstruction. Our framework is anchored by two principal innovations. 
First, Multi-Granularity Value Guidance Learning (MVGL) module learns the value map with multi-granularity guidance, coupling an implicit self-similarity statistic with an explicit DINOv2 prior. This design is used to generate a value map to guide holistic content reordering. 
Second, the novel dual-path Mamba–Conv Mixer (MCM) Block is designed to harmonize value-driven global sequence modeling with adaptive local feature extraction, thereby effectively addressing the distinct statistical characteristics of heterogeneous underwater scenes. 
A Cross-Feature Bridge (CFB) performs context-conditioned channel–spatial fusion on the skip connections to stabilize color correction while preserving detail.
Extensive experimental validation demonstrates that the proposed method achieves state-of-the-art performance across multiple underwater image benchmarks, surpassing wMamba by 0.89 dB in PSNR. Meanwhile, the framework provides enhanced explainability via its value-aware processing mechanism. 

\bibliographystyle{IEEEtran}
\bibliography{ref}

\vfill

\end{document}